\documentclass{article} 
\usepackage{iclr2021_conference,times}
\usepackage{algorithm,algorithmic,amsmath}
\usepackage{wrapfig}

\usepackage{amsmath,amsfonts,bm}

\def\eqref#1{equation~\ref{#1}}

\def\1{\bm{1}}

\DeclareMathAlphabet{\mathsfit}{\encodingdefault}{\sfdefault}{m}{sl}
\SetMathAlphabet{\mathsfit}{bold}{\encodingdefault}{\sfdefault}{bx}{n}

\usepackage{hyperref}
\usepackage{url}
\usepackage{graphicx}
\usepackage{subcaption}

\title{Gradient-Masked Federated 
Optimization}

\author
{Irene Tenison$^{1,2}$, Sreya Francis$^{1,2}$, Irina Rish$^{1,2}$\\
\normalsize{MILA$^{1}$},
\normalsize{University of Montreal$^{2}$}
}

\iclrfinalcopy 
\begin{document}

\maketitle

\begin{abstract}
Federated Averaging (FedAVG) has become the most popular federated learning algorithm due to its simplicity and low communication overhead.  We use simple examples to show that FedAVG has the tendency to sew together the optima across the participating clients.  These sewed optima exhibit poor generalization when used on a new client with new data distribution. Inspired by the invariance principles in \citep{arjovsky2019invariant,parascandolo2020learning}, we focus on learning a model that is locally optimal across the different clients simultaneously. We propose a modification to FedAVG algorithm to include masked gradients (AND-mask from \citep{parascandolo2020learning})  across the clients and uses them to carry out an additional server model update.  We show that this algorithm achieves better accuracy (out-of-distribution) than FedAVG, especially when the data is non-identically distributed across clients.

\end{abstract}

\section{Introduction}

Federated learning \citep{McMahan2017CommunicationEfficientLO, DBLP:journals/corr/KonecnyMYRSB16,DBLP:journals/corr/abs-1912-04977,DBLP:conf/nips/WangLLJP20} has emerged as an attractive distributed learning paradigm where a large part of the model training is pushed towards edge devices. Unlike traditional centralized learning, federated learning enables collective training of a global model (or server model) for a common objective while the training data remains distributed at the clients' end. This ensures data privacy to an extent and helps in training a shared model without heavy computational or storage requirements at the server.

There are several callenges to federated optimization. The key challenges being 1) slow and unreliable network connection between the client and the server \citep{pmlr-v119-karimireddy20a}, 2) heterogenity among client devices and device usage pattern, 3) non-IID (non-independent and identically distributed) data at clients due to varying client behaviour. The most popular algorithm in federated setting emphasising on handling non-iid and unbalanced data and communication bottleneck is FedAVG \citep{McMahan2017CommunicationEfficientLO}. It performs multiple local steps at the client before aggregating the parameters for the server model. While it has shown success in several areas including tackling communication overhead, its performance on non-IID data is an active area of research \citep{haddadpour2019convergence, li2020federated, wang2020tackling,pmlr-v119-karimireddy20a}. Extension of this research to out-of-distribution generalization in a federated setting is relatively new and little explored.

In this paper, we show that averaging parameters of local client models as in FedAVG, results in a server model having minima close to locations of global minima of the individual client models, instead of at the consistenet minima; and with a loss value equal to the average of client losses at that location using the concept of loss surfaces. The same have been described theoretically as "gradient bias" in \citep{yao2020federated} and as "client-drift" in \citep{pmlr-v119-karimireddy20a}. Though the server model in a federated setting aims to learn a global objective, averaging of parameters fails in effectively doing this since the minima (both global and local) of each client varies drastically due to dissimilar user behaviour. This is more prominent when the data is non-iid. To handle this, we introduce a modification to the  FedAVG algorithm. FedGMA (Federated Gradient Masked Averaging) use a AND-Masked \citep{parascandolo2020learning} gradient update along with parameter averaging (as in FedAVG) to ensure update steps in the direction of the optimal minima across clients.

\section{Convergence of FedAVG}

 \begin{figure}[t]
\begin{subfigure}{.33\textwidth}
  \centering
  \includegraphics[width=1\linewidth]{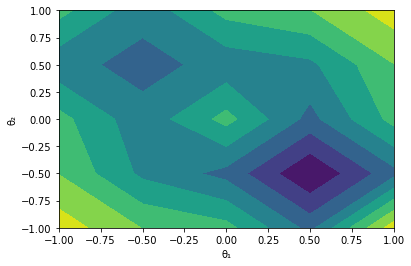}  
  \caption{Loss surface of client A}
  \label{fig:figa}
\end{subfigure}
\begin{subfigure}{.33\textwidth}
  \centering
  \includegraphics[width=1\linewidth]{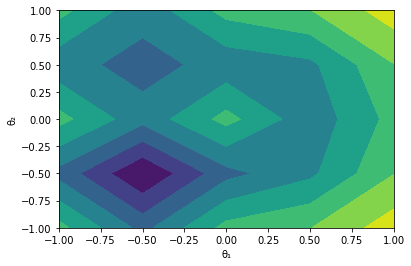}  
  \caption{Loss surface of client B}
  \label{fig:fig6}
\end{subfigure}
\begin{subfigure}{.33\textwidth}
  \centering
  \includegraphics[width=1\linewidth]{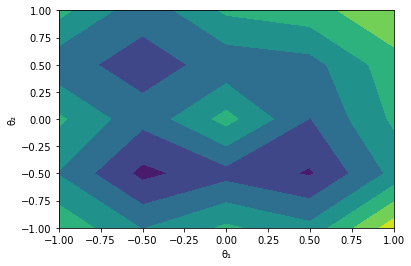}  
  \caption{FedAVG server loss surface}
  \label{fig:fig7}
\end{subfigure}

\caption{Loss surfaces of two clients, A and B, with respect to a two-parameter vector ($\theta = (\theta_1,\theta_2)$) varied across their limits, and their average loss surface representing FedAVG server loss surface. Algorithms that average client model parameters or gradients with respect to the client model parameters for server model update forgoes information that can identify shared patterns or consistent minima across clients.}
\label{fig:figx}
\end{figure}

Consider figures 1a. and 1b., showing the top view of loss surfaces of two fictional clients, A and B respectively. Note that client A has a local minima at (-0.5, 0.5) (here after referred to as $L_{min}$) and a global minima at (0.5, -0.5) (here after referred to as $G_{A_{min}}$) and client B has a local minima at (-0.5, 0.5) ($L_{min}$) and a global minima at (-0.5, -0.5) (here after referred to as $G_{B_{min}}$). Depending on the initial values of the parameters, models trained on a client's local data converges to either of its two minima. It is observable that the global minima of the two clients ($G_{A_{min}}$ and $G_{B_{min}}$) are apart and converging to either one of them would result in a high loss on data from the other client. Whereas, the local minima ($L_{min}$) of both the clients are overlapping and the loss of a client corresponding to the data from the other client will be less for this local minima. Hence, a server model converging to this local minima ($L_{min}$) will have better generalization on other clients. This is backed by the principle that good explanations are hard to vary \citep{parascandolo2020learning}.

In FedAVG, where the client model parameters are averaged to obtain the server model, the server loss can be  approximated to the average of the loss values of clients participating in the update (server loss with MSE and cross-entropy loss at clients approximated in Appendix A.1) and is as shown in Figure 1c. It is observable that the approximated server loss is minimum at $G_{A_{min}}$ and $G_{B_{min}}$, though generalization would be better at the shared or consistent local minima $L_{min}$. FedAVG optimizes to a solution that sew together the individual objectives of the clients, instead of identifying the invarient and consistent patterns across the clients.

\section{Proposed Algorithm - FedGMA}

In this section, we describe the proposed algorithm, FedGMA(Federated Gradient Masked Averaging) in detail. The algorithm has three main steps: 1) local updates  to the client model, 2) aggregation of and-masked gradients with respect to the initial parameters of the respective client model, 3) server model updates where each parameter is the average of individual client parameters along with an update step along the masked gradient.

Similar to FedAVG, the proposed algorithm maintains a shared server model, aiming at learning a common objective, and a local model for each client, trained on their local data (from initial state parameters equal to the server model parameters at that communication round), as shown in algorithm 1. In our setting, we have assumed that each client undergoes the same number($n$) of local stochastic gradient descent steps. Performance of FedGMA corresponding to the number of local epochs will be studied later. At the end of $n$ epochs, each client returns the gradient of their loss with respect to the initial parameters of the respective client model and the client model parameters. These gradients are masked at the server using an and-mask and the mean of the masked gradients is considered for the server model update along with the weighted average of the client model parameters. 

Seperate step sizes were used for the server update($\eta_s$ as in line 9 of ServerUpdate in algorithm 1) and client update($\eta_c$ as in line 4 of ClientUpdate in algorithm 1). Larger server model step size enables faster convergence of the server model while a smaller client model step size keeps the client models from drifting farther away from the server model, thereby preventing each client model from converging to their global minima \citep{pmlr-v119-karimireddy20a}. 

\begin{algorithm}[t]
\caption{Gradient Masked Averaging }
$\textbf{ServerUpdate:}
  $
  \begin{algorithmic}[1]
    \STATE $\text{Initialize }w_0
    $
    \FOR{$\text{each server epoch, t = 1,2,3,... }$}
    \STATE$\text{Choose C clients at random}$
      \FOR{$\text{each client, k}$}
        \STATE $grad^k,w^k = ClientUpdate(w_{t-1})
        $
      \ENDFOR
      \STATE $grad = \frac{m}{n_k} \sum_k \frac{1}{m_k} Mask(grad^k)$
      \STATE $w = \frac{m}{n_k} \sum_k \frac{1}{m_k} w^k$
      \STATE $w_t = w - \eta_s\text{ } grad
      $
    \ENDFOR
    
  \end{algorithmic}
  
  $\textbf{ClientUpdate(w):}
  $
  \begin{algorithmic}[1]
    \STATE $\text{Initialize }w_0 = w
    $
    \FOR{$\text{each local client epoch, i=0,1,2,3,..,n}$}
    \STATE $g_i = \nabla_{w_i} L(w_i)$
      \STATE $w_{i+1} = w_{i} - \eta_c\text{ } g_i
      $
    \ENDFOR
    \RETURN  $g_i,w_{i+1}$ $\text{ to server}$
  \end{algorithmic}
\end{algorithm}

\textbf{AND-Mask.} Inspired by the work \citep{parascandolo2020learning}, we use an AND-Mask on gradients($grad^k$) before averaging (weighted with respect to the number of examples in that client \citep{McMahan2017CommunicationEfficientLO}) them for server update. The AND-Mask is mathematically similar to a 'logical and'. Intuitively, it can be visualized as a zeroing out the gradient components that have inconsistent signs across clients. Formally, masked gradients vanishes for components where there are less than $t\in {k/2+1,...,k}$ (where k is the total number of participating clients) gradient components that agree to the same sign across clients, and is one otherwise. That is, 50\% or more gradient components should be having agreeing gradient signs across clients for the gradient to be activated and included in the server update. This ensures that the server model converges to a solution consistent across clients. AND-Mask follows the strategy that to converge to a consistent minima that generalises better, only consistent gradient directions across clients corresponding to a "logical AND" are to be pursued. Similar to AND-Mask implementation in \citep{parascandolo2020learning} for environments sharing invariant mechanisms, AND-Masked gradient component with respect to a client $i$ in a federated setting can be represented as $mask(g_i) = g_i * I_{\{pk \leq | \sum_k \text{sign}(g_{c\in k}) |\}}$ where, $p \in [0,1]$ is the percentage of total number of clients, $k$, whose gradients with respect to client parameters need to be consistent and $I$ is the boolean indicator function facilitating this. Averaging masked gradients is equivalent to averaging the gradients directly when all the client gradients being considered are of the same sign. AND-mask can also be viewed as a regularizer which prevents overfitting to certain clients by prohibiting further updates in model parameters when the gradients with respect to the parameters are inconsistent across clients.

\section{Experiments}

We run experiments on non-convex optimization represented by a CNN for multi-class digit recognition and binary classification (less than 5 and greater than or equal to 5) on colored MNIST, where the data was distributed both identically and non-identically across clients and tested on an out-of-distribution dataset.
Inspired by the experiments in \citep{arjovsky2019invariant}, for binary classification, the data in all train environments or clients were colored red if the digit was less than 5 and green otherwise with probabilities varying between 0.1 and 0.2 for different clients. For the out-of-distribution test set, the coloring schema was reversed (green if the digit was less than 5 and red otherwise) with a probability of 0.9 (see Appendix for samples). For multi-class digit recognition, inspired by the experiments in \citep{arpit2019predicting}, each class or digit was assigned two background colors and two foreground colors that were unique to the class. For the out-of-distribution test images, foreground and background colors were chosen randomly from a set of 10 set of colors independent of the training colors (see Appendix for samples). The distribution of data across clients was done in accordance to the experiments in \citep{McMahan2017CommunicationEfficientLO}. For IID distribution, the entire train data was shuffled and split among the clients such that every client would get images of all digits. For Non-IID distribution, the train data was split such that each client would get approximately equal number of images from exactly two digits.

In the experiments, we have chosen the AND-Mask threshold percentage hyperparameter, $p = 0.8$, that is, atleast 80\% of the clients should be agreeing on gradient signs across clients for the gradients to not be zeroed out. Server step size (lr\_s in Figure 2) is another hyperparameter introduced in this algorithm. We have considered FedAVG as the baseline and FedGMA was studied with the same hyperparameters as the FedAVG baseline for effective comparison. This includes a 3-epoch client training with SGD(Stochastic Gradient Descent) optimizer having a step size of 0.001. It was observed that the proposed algorithm outperforms FedAVG in generalization to out-of-distribution test sets for both IID and Non-IID settings, though the performance improvement was more prominent when the data distribution among clients was non-IID.

 \begin{figure}[t]
 \centering
\begin{subfigure}{.4\textwidth}
  \centering
  \includegraphics[width=1\linewidth]{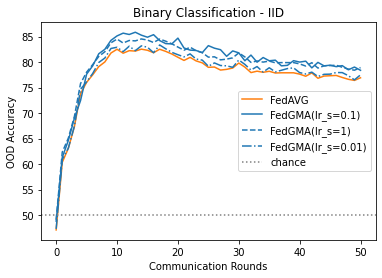}  
  \caption{}
  \label{fig:figb}
\end{subfigure}
\begin{subfigure}{.4\textwidth}
  \centering
  \includegraphics[width=1\linewidth]{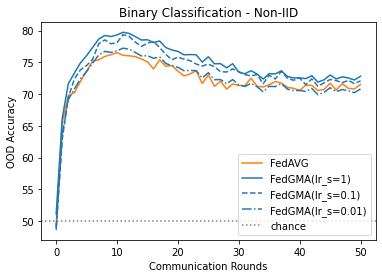}  
  \caption{}
  \label{fig:fig1}
\end{subfigure}
\begin{subfigure}{.4\textwidth}
  \centering
  \includegraphics[width=1\linewidth]{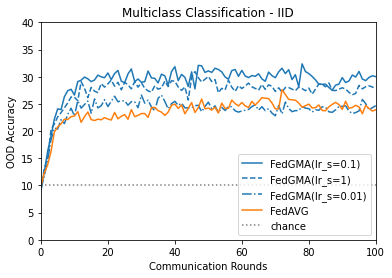}  
  \caption{}
  \label{fig:fig2}
\end{subfigure}
\begin{subfigure}{.4\textwidth}
  \centering
  \includegraphics[width=1\linewidth]{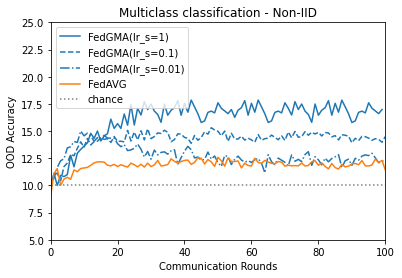}  
  \caption{}
  \label{fig:fig3}
\end{subfigure}

\caption{Out-of-distribution (OOD) accuracy vs. communication rounds plots of binary classification in IID (a) and non-IID (b) settings and multiclass classification in IID (c) and non-IID (d) settings with different server step size(lr\_s) values. FedGMA had a better OOD generalization performance than FedAVG. It is more prominent with a significant improvement in OOD accuracy when data is distributed non-identically for both binary and multiclass classification with the right server step size hyperparameter.}
\label{fig:fig8}
\end{figure}

\section{Conclusion and Future Work}

In this paper, we show how FedAVG and other federated optimization algorithms that average gradients to update the server model, converges to a minima that is a sewn together version of all client minima. We introduce a modification to the client update step of FedAVG to incorporate an update step using AND-Mask on gradients with respect to initial client model parameters. This step ensures that the server model converges to an inconsistent minima across clients. We experimentally show that FedGMA achieves a better generalization performance on out-of-distribution test images than FedAVG.

In the future, we would like to expand our work to include more experiments on other state-of-the art datasets like CIFAR-10 and Shakespeare language dataset and comparisons with other existing federated optimization algorithms like SCAFFOLD and FedProx. We would like to study the effect of hyperparameters like $t$, the threshold for the number of client gradients to be consistent, and $n$, the number of local epochs.
Understanding the effect of the AND-Mask on existing federated optimization algorithms would be an interesting area to explore as well. 

\subsection*{Acknowledgment}
We would like to thank Karthik Ahuja for his feedbacks and reviews.

\bibliography{iclr2021_conference}

\begin{thebibliography}{12}
\providecommand{\natexlab}[1]{#1}
\providecommand{\url}[1]{\texttt{#1}}
\expandafter\ifx\csname urlstyle\endcsname\relax
  \providecommand{\doi}[1]{doi: #1}\else
  \providecommand{\doi}{doi: \begingroup \urlstyle{rm}\Url}\fi

\bibitem[Arjovsky et~al.(2019)Arjovsky, Bottou, Gulrajani, and
  Lopez-Paz]{arjovsky2019invariant}
Martin Arjovsky, L{\'e}on Bottou, Ishaan Gulrajani, and David Lopez-Paz.
\newblock Invariant risk minimization.
\newblock \emph{arXiv preprint arXiv:1907.02893}, 2019.

\bibitem[Arpit et~al.(2019)Arpit, Xiong, and Socher]{arpit2019predicting}
Devansh Arpit, Caiming Xiong, and Richard Socher.
\newblock Predicting with high correlation features, 2019.

\bibitem[Haddadpour \& Mahdavi(2019)Haddadpour and
  Mahdavi]{haddadpour2019convergence}
Farzin Haddadpour and Mehrdad Mahdavi.
\newblock On the convergence of local descent methods in federated learning,
  2019.

\bibitem[Kairouz et~al.(2019)Kairouz, McMahan, Avent, Bellet, Bennis, Bhagoji,
  Bonawitz, Charles, Cormode, Cummings, D'Oliveira, Rouayheb, Evans, Gardner,
  Garrett, Gasc{\'{o}}n, Ghazi, Gibbons, Gruteser, Harchaoui, He, He, Huo,
  Hutchinson, Hsu, Jaggi, Javidi, Joshi, Khodak, Konecn{\'{y}}, Korolova,
  Koushanfar, Koyejo, Lepoint, Liu, Mittal, Mohri, Nock, {\"{O}}zg{\"{u}}r,
  Pagh, Raykova, Qi, Ramage, Raskar, Song, Song, Stich, Sun, Suresh,
  Tram{\`{e}}r, Vepakomma, Wang, Xiong, Xu, Yang, Yu, Yu, and
  Zhao]{DBLP:journals/corr/abs-1912-04977}
Peter Kairouz, H.~Brendan McMahan, Brendan Avent, Aur{\'{e}}lien Bellet, Mehdi
  Bennis, Arjun~Nitin Bhagoji, Keith Bonawitz, Zachary Charles, Graham Cormode,
  Rachel Cummings, Rafael G.~L. D'Oliveira, Salim~El Rouayheb, David Evans,
  Josh Gardner, Zachary Garrett, Adri{\`{a}} Gasc{\'{o}}n, Badih Ghazi,
  Phillip~B. Gibbons, Marco Gruteser, Za{\"{\i}}d Harchaoui, Chaoyang He, Lie
  He, Zhouyuan Huo, Ben Hutchinson, Justin Hsu, Martin Jaggi, Tara Javidi,
  Gauri Joshi, Mikhail Khodak, Jakub Konecn{\'{y}}, Aleksandra Korolova,
  Farinaz Koushanfar, Sanmi Koyejo, Tancr{\`{e}}de Lepoint, Yang Liu, Prateek
  Mittal, Mehryar Mohri, Richard Nock, Ayfer {\"{O}}zg{\"{u}}r, Rasmus Pagh,
  Mariana Raykova, Hang Qi, Daniel Ramage, Ramesh Raskar, Dawn Song, Weikang
  Song, Sebastian~U. Stich, Ziteng Sun, Ananda~Theertha Suresh, Florian
  Tram{\`{e}}r, Praneeth Vepakomma, Jianyu Wang, Li~Xiong, Zheng Xu, Qiang
  Yang, Felix~X. Yu, Han Yu, and Sen Zhao.
\newblock Advances and open problems in federated learning.
\newblock \emph{CoRR}, abs/1912.04977, 2019.
\newblock URL \url{http://arxiv.org/abs/1912.04977}.

\bibitem[Karimireddy et~al.(2020)Karimireddy, Kale, Mohri, Reddi, Stich, and
  Suresh]{pmlr-v119-karimireddy20a}
Sai~Praneeth Karimireddy, Satyen Kale, Mehryar Mohri, Sashank Reddi, Sebastian
  Stich, and Ananda~Theertha Suresh.
\newblock {SCAFFOLD}: Stochastic controlled averaging for federated learning.
\newblock In Hal~Daumé III and Aarti Singh (eds.), \emph{Proceedings of the
  37th International Conference on Machine Learning}, volume 119 of
  \emph{Proceedings of Machine Learning Research}, pp.\  5132--5143. PMLR,
  13--18 Jul 2020.
\newblock URL \url{http://proceedings.mlr.press/v119/karimireddy20a.html}.

\bibitem[Konecn{\'{y}} et~al.(2016)Konecn{\'{y}}, McMahan, Yu, Richt{\'{a}}rik,
  Suresh, and Bacon]{DBLP:journals/corr/KonecnyMYRSB16}
Jakub Konecn{\'{y}}, H.~Brendan McMahan, Felix~X. Yu, Peter Richt{\'{a}}rik,
  Ananda~Theertha Suresh, and Dave Bacon.
\newblock Federated learning: Strategies for improving communication
  efficiency.
\newblock \emph{CoRR}, abs/1610.05492, 2016.
\newblock URL \url{http://arxiv.org/abs/1610.05492}.

\bibitem[Li et~al.(2020)Li, Sahu, Zaheer, Sanjabi, Talwalkar, and
  Smith]{li2020federated}
Tian Li, Anit~Kumar Sahu, Manzil Zaheer, Maziar Sanjabi, Ameet Talwalkar, and
  Virginia Smith.
\newblock Federated optimization in heterogeneous networks, 2020.

\bibitem[McMahan et~al.(2017)McMahan, Moore, Ramage, Hampson, and
  y~Arcas]{McMahan2017CommunicationEfficientLO}
H.~McMahan, Eider Moore, D.~Ramage, S.~Hampson, and Blaise~Ag{\"u}era y~Arcas.
\newblock Communication-efficient learning of deep networks from decentralized
  data.
\newblock In \emph{AISTATS}, 2017.

\bibitem[Parascandolo et~al.(2020)Parascandolo, Neitz, Orvieto, Gresele, and
  Schölkopf]{parascandolo2020learning}
Giambattista Parascandolo, Alexander Neitz, Antonio Orvieto, Luigi Gresele, and
  Bernhard Schölkopf.
\newblock Learning explanations that are hard to vary, 2020.

\bibitem[Wang et~al.(2020{\natexlab{a}})Wang, Liu, Liang, Joshi, and
  Poor]{DBLP:conf/nips/WangLLJP20}
Jianyu Wang, Qinghua Liu, Hao Liang, Gauri Joshi, and H.~Vincent Poor.
\newblock Tackling the objective inconsistency problem in heterogeneous
  federated optimization.
\newblock In Hugo Larochelle, Marc'Aurelio Ranzato, Raia Hadsell,
  Maria{-}Florina Balcan, and Hsuan{-}Tien Lin (eds.), \emph{Advances in Neural
  Information Processing Systems 33: Annual Conference on Neural Information
  Processing Systems 2020, NeurIPS 2020, December 6-12, 2020, virtual},
  2020{\natexlab{a}}.
\newblock URL
  \url{https://proceedings.neurips.cc/paper/2020/hash/564127c03caab942e503ee6f810f54fd-Abstract.html}.

\bibitem[Wang et~al.(2020{\natexlab{b}})Wang, Liu, Liang, Joshi, and
  Poor]{wang2020tackling}
Jianyu Wang, Qinghua Liu, Hao Liang, Gauri Joshi, and H.~Vincent Poor.
\newblock Tackling the objective inconsistency problem in heterogeneous
  federated optimization, 2020{\natexlab{b}}.

\bibitem[Yao et~al.(2020)Yao, Huang, Zhang, Li, and Sun]{yao2020federated}
Xin Yao, Tianchi Huang, Rui-Xiao Zhang, Ruiyu Li, and Lifeng Sun.
\newblock Federated learning with unbiased gradient aggregation and
  controllable meta updating, 2020.

\end{thebibliography}
\bibliographystyle{iclr2021_conference}

\appendix
\section{Appendix}

\subsection{Approximate Server Loss}
When loss at the client is MSE Loss,\\
\small
\begin{equation*}
\begin{split}
\text{Loss at client A, } J_A &= (y-w_Ax)^2 \\
\text{Loss at client B, } J_B &= (y-w_Bx)^2 \\
\text{Average loss surface of the two clients} & = \frac{1}{2} (J_A+J_B) \\ &= \frac{1}{2} (({w_A}^2+{w_B}^2)x^2+2y^2-2(w_A+w_B)xy) \\
&= \frac{1}{2} ({w_A}^2+{w_B}^2)x^2 + y^2 -  (w_A+w_B)xy  \\
\text{Loss at FedAVG server, } J &= (y-wx)^2 \text{ where, } w = \frac{w_A+w_B}{2}\\
&= (\frac{w_A+w_B}{2})^2x^2 + y^2 - (w_A+w_B)xy \\
&= \frac{1}{2} (J_A+J_B) - \frac{1}{2} ({w_A}^2+{w_B}^2)x^2 + (\frac{w_A+w_B}{2})^2x^2\\
& \approx \frac{1}{2} (J_A+J_B) 
\end{split}
\end{equation*}
\normalsize
When the weights are close to zero or the variance between the two clients are small, $(\frac{w_A+w_B}{2})^2x^2 - \frac{1}{2} ({w_A}^2+{w_B}^2)x^2$ becomes negligible and $J\approx \frac{1}{2} (J_A+J_B)$.\\
\vspace{1cm}

When loss at the client is cross entropy Loss,\\
\small
\begin{equation*}
\begin{split}
\text{Loss at client A, } J_A &= \sum y \log (w_Ax) \\
\text{Loss at client B, } J_B &= \sum y \log (w_Bx) \\
\text{Average loss surface of the two clients} & = \frac{1}{2} (J_A+J_B) \\ &= \frac{1}{2} (\sum y \log (w_Ax) + \sum y \log (w_Bx)) \\
&= \frac{1}{2} (\sum y \log (w_A) + \sum y \log (w_B)) + \sum y \log x\\
&= \frac{1}{2} (\sum y \log (w_Aw_B) + \sum y \log x\\
&= \sum y \log (\sqrt{w_Aw_B}) + \sum y \log x \\
\text{Loss at FedAVG server, } J &= \sum y \log (wx) \text{ where, } w = \frac{w_A+w_B}{2}\\
&=\sum y \log (\frac{w_A+w_B}{2}x)\\
&= \sum y \log (\frac{w_A+w_B}{2}) + \sum y \log x 
\end{split}
\end{equation*}
\normalsize
When $\frac{w_A+w_B}{2} \approx \sqrt{w_Aw_B}$, $J\approx \frac{1}{2} (J_A+J_B)$. This happens when the weights are all very small and close to zero.

\subsection{Data Samples - CMNIST}

\begin{figure}[H]
\centering
\begin{subfigure}{.4\textwidth}
  \centering
  \includegraphics[width=1\linewidth]{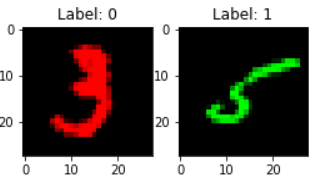}  
  \caption{}
  \label{fig:figc}
\end{subfigure}
\begin{subfigure}{.4\textwidth}
  \centering
  \includegraphics[width=1\linewidth]{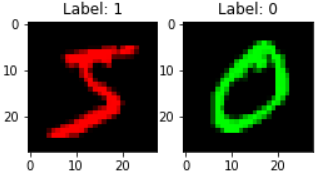}  
  \caption{}
  \label{fig:fig4}
\end{subfigure}
\caption{Sample images from colored MNIST for binary classification. a) In train environments, digits less than 5 marked by label 0 are colored red, while the digits greater than or equal to 5, marked by label 1 are colored green. b) In the test environment, the colors are reversed. That is, digits of label 0 are colored green and digits of label 1 are colored red. \citep{arjovsky2019invariant}}
\end{figure}

\begin{figure}[H]
\centering
\begin{subfigure}{.3\textwidth}
  \centering
  \includegraphics[width=1\linewidth]{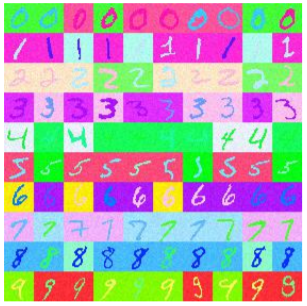}  
  \caption{}
  \label{fig:figd}
\end{subfigure}
\begin{subfigure}{.3\textwidth}
  \centering
  \includegraphics[width=1\linewidth]{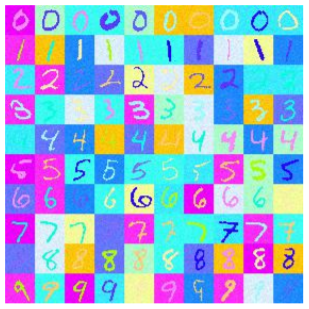}  
  \caption{}
  \label{fig:fig5}
\end{subfigure}
\caption{Samples images colored MNIST for multi-class classification. a) In train environments, each class is given two unique foreground colors and two unique background colors. b) In the test enviroment, the foreground color and the background color of each image is chosen randomly from the list of colors independent of the colors in the train environments. \citep{arpit2019predicting}}
\end{figure}
\end{document}